\documentclass[10pt,twocolumn,letterpaper]{article}

\usepackage{btas}
\usepackage{graphicx,tabularx}
\usepackage{amsmath,amssymb}
\usepackage{cmap}
\usepackage{txfonts}
\usepackage{framed,multirow,units,enumitem,url,caption,subcaption,booktabs}
\usepackage[misc]{ifsym}
\usepackage{microtype}
\usepackage{indentfirst}
\usepackage{fancyhdr}

\newcommand{\gC}{\ensuremath{C}} 
\newcommand{\gCLAS}{\ensuremath{\mathcal{I}}} 
\newcommand{\gCLASc}[1]{\ensuremath{\gCLAS_{#1}}} 
\newcommand{\gCLUS}{\ensuremath{\mathcal{C}}} 
\newcommand{\gCLUSc}[1]{\ensuremath{\gCLUS_{#1}}} 
\newcommand{\gDELTAccH}[2]{\ensuremath{\gH{\delta}\left(#1,#2\right)}} 
\newcommand{\gG}{\ensuremath{\mathcal{G}}} 
\newcommand{\gGn}[1]{\ensuremath{\gB{g}_{#1}}} 
\newcommand{\gGnH}[1]{\ensuremath{\gH{\gB{g}}_{#1}}} 
\newcommand{\gLAMBDAn}[1]{\ensuremath{\ell_{#1}}} 
\newcommand{\gM}{\ensuremath{\mu}} 
\newcommand{\gMcH}[1]{\ensuremath{\gH{\gM}_{#1}}} 
\newcommand{\gN}{\ensuremath{N}} 
\newcommand{\gNc}[1]{\ensuremath{\gN_{#1}}} 

\newcommand{\gB}[1]{\ensuremath{\mathbf{#1}}} 
\newcommand{\gH}[1]{\ensuremath{\widehat{#1}}} 
\newcommand{\gL}[1]{\ensuremath{{#1}_L}} 
\newcommand{\gE}[1]{\ensuremath{{#1}_E}} 

\newcommand{\joint}[1]{\texttt{#1}}
\newcommand{\DBI}{\ensuremath{\mathrm{DBI}}\xspace}
\newcommand{\SC}{\ensuremath{\mathrm{SC}}\xspace}
\newcommand{\ROC}{\ensuremath{\mathrm{ROC}}\xspace}
\newcommand{\PR}{\ensuremath{\mathrm{PR}}\xspace}
\newcommand{\PURITY}{\ensuremath{\mathrm{P}}\xspace}
\newcommand{\RI}{\ensuremath{\mathrm{RI}}\xspace}
\newcommand{\F}{\ensuremath{\mathrm{F}}\xspace}
\newcommand{\JI}{\ensuremath{\mathrm{JI}}\xspace}
\newcommand{\FMI}{\ensuremath{\mathrm{FMI}}\xspace}
\newcommand{\TP}{\ensuremath{\mathrm{TP}}\xspace}
\newcommand{\FP}{\ensuremath{\mathrm{FP}}\xspace}
\newcommand{\FN}{\ensuremath{\mathrm{FN}}\xspace}
\newcommand{\TN}{\ensuremath{\mathrm{TN}}\xspace}

\newcommand{\new}{\ensuremath{\mathrm{new}}\xspace}
\newcommand{\old}{\ensuremath{\mathrm{old}}\xspace}

\usepackage[breaklinks=true,colorlinks,bookmarks=false]{hyperref}

\btasfinalcopy 

\ifbtasfinal\fi 

\begin{document}

\title{
You~Are~How~You~Walk:
Uncooperative~MoCap~Gait~Identification
for~Video~Surveillance
with~Incomplete~and~Noisy~Data
}

\author{
Michal Balazia\\
{\tt\small xbalazia@mail.muni.cz}
\and
Petr Sojka\\
{\tt\small sojka@fi.muni.cz}
\and
Faculty of Informatics, Masaryk University, Botanick\'a 68a, 602\,00 Brno, Czech Republic
}

\maketitle

\pagestyle{plain}
\thispagestyle{fancy}
\fancyhead[C]{3rd IEEE/IAPR International Joint Conference on Biometrics 2017, preprint}

\begin{abstract}
This work offers a design of a video surveillance system based on a soft biometric -- gait identification from MoCap data. The main focus is on two substantial issues of the video surveillance scenario: (1)~the walkers do not cooperate in providing learning data to establish their identities and (2)~the data are often noisy or incomplete. We show that only a few examples of human gait cycles are required to learn a projection of raw MoCap data onto a low-dimensional sub-space where the identities are well separable. Latent features learned by Maximum Margin Criterion (MMC) method discriminate better than any collection of geometric features. The MMC method is also highly robust to noisy data and works properly even with only a fraction of joints tracked. The overall workflow of the design is directly applicable for a day-to-day operation based on the available MoCap technology and algorithms for gait analysis. In the concept we introduce, a walker's identity is represented by a cluster of gait data collected at their incidents within the surveillance system: They are how they walk.
\end{abstract}

\section{Introduction}
\label{intro}

Public safety issues are constantly evolving and security monitoring agencies are facing more challenges than ever before. Offering a range of security systems indispensable to investigators, the surveillance industry appears to be at the beginning of a massive expansion. Video surveillance technology records video footage for the potential future identification of suspicious individuals and activities. Many public places, such as banks and airports, already have surveillance cameras installed, but these require intelligent approaches to human identification. A useful early-warning system would analyze the collected video footage and release an alert before an adverse event takes place. Triggered by detection of an abnormal behavior, the system would instantly identify all participants in the scene, rapidly investigate their previous activities, and launch the tracking of the suspects.

A typical video surveillance environment has some crucial characteristics that need to be taken into account when designing an identification system. Data are captured by a system of video cameras covering a large tracking space. People walk in various directions, at various speeds, and are often in crowds. They wear a variety of clothes and shoes and often carry large objects. Since they do not cooperate in this, an eventual learning model is only available in an externally annotated database. A central database stores hundreds of subject identities that are encountered repeatedly, each contributing with multiple biometric samples. Identification has to be performed in real time, that is, in a few seconds. Tracking and identification have to be automatic as operator interventions are slow and costly.

Gait (walk) pattern has several attractive properties as a soft biometric trait. From the surveillance perspective, gait pattern biometrics is appealing for its possibility of being performed at a distance and without body-invasive equipment or subject cooperation. This enables sample acquisition even without a subject's consent. The goal of this work is to design a method for identifying individuals from videos from their gait pattern.

Uncooperative gait identification has been addressed by Martin-Felez and Xiang~\cite{MX14} by casting gait identification as a bipartite ranking problem. Their model learns a ranking function in a higher dimensional space where true matches and wrong matches become more separable than in the original space. The output of the model is a ranking function which gives a higher score if a pair of gait templates belong to the same person than to different people.

Chen and Xu~\cite{CX16} extend the bipartite ranking approach by integrating sparse coding re-ranking and multi-view hypergraph-based re-ranking framework, calling it a sparse coding multi-view hypergraph learning re-ranking method.

\section{Identification Workflow}
\label{flow}

In accordance with the outlined video surveillance environment, our human identification system has the following 4-phase workflow (see Figure~\ref{f1}):

\begin{figure*}[tb]
\centering
\includegraphics[width=\textwidth]{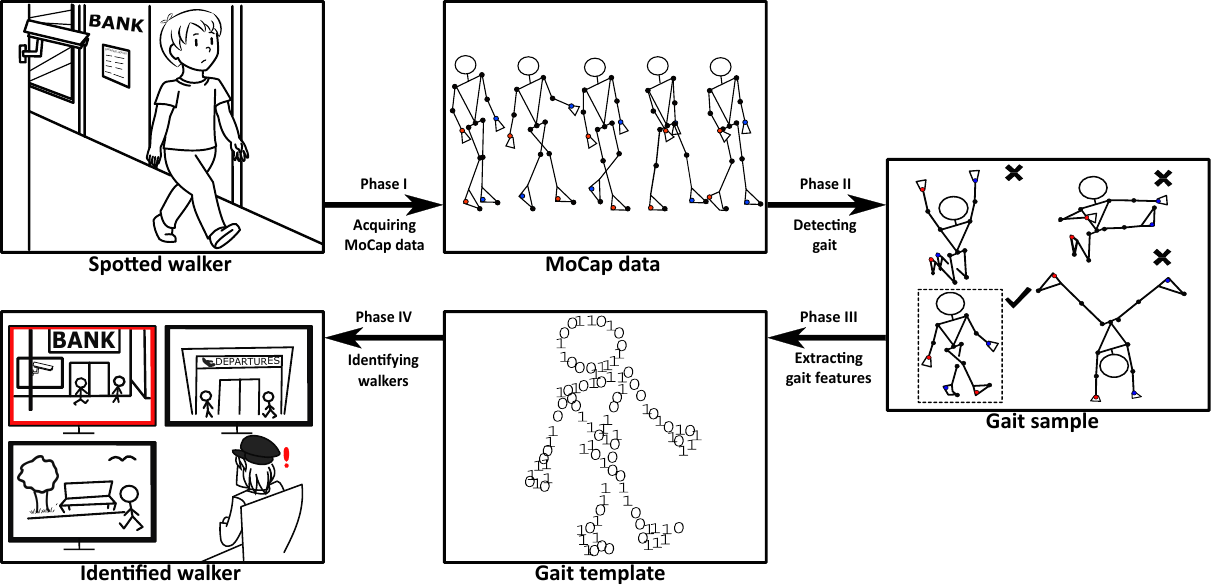}
\caption{Video surveillance workflow. A person is (I)~captured on an RGB-D camera in the form of MoCap data, (II)~gait is then detected to form a gait sample, from which (III)~gait template is extracted and (IV)~the person is identified.}
\label{f1}
\end{figure*}

\subsection*{Phase I -- Acquiring Motion Capture Data}
Motion capture (MoCap) technology acquires video clips of people and derives structural motion data. The format maintains an overall structure of the human body and holds the estimated 3D positions of the main anatomical landmarks as the person moves. MoCap data can be collected by RGB-D sensors such as Microsoft Kinect, Asus Xtion or Vicon. For a schematic visualization, a simplified stick figure representing the human skeleton (a graph of joints connected by bones) can be automatically recovered from the values of body point spatial coordinates. With recent rapid improvements in sensor technology and pose estimation techniques~\cite{DF16,HPLAYL16}, an accurate and affordable MoCap system~\cite{HALP13} is at our disposal to aid gait identification for applications in video surveillance.

\subsection*{Phase II -- Detecting Gait Cycles}

People spotted in our tracking space do not walk all the time; on the contrary, they perform various activities. Identifying people from gait requires processing video segments where they are actually walking. Gait cycles need to be first filtered out from the video sequences of general motion. There are methods~\cite{AMAMR15,VSBZ12} for detecting gait cycles directly as well as action recognition methods~\cite{CCH12,HCCCLW15,KN14,LLMYD14,VR14} that need a demonstrative example of a gait cycle to query general motion sequences.

\subsection*{Phase III -- Extracting Gait Features}

Once a query motion clip has been cut into clean gait cycles, the identification mechanism proceeds with transforming the sample of raw MoCap data into a representation that contains discriminative gait information. A collection of extracted gait features builds a gait template which serves as the walker's signature. But as in video surveillance environment the walkers cannot be relied upon to be cooperative, we are left with the problem of obtaining highly discriminative gait features for the walkers without a labeled learning dataset containing their very own samples.

Many research groups investigate the discriminatory power of geometric gait features designed by hand and without any statistical learning. They typically combine static body parameters (bone lengths, person's height) with dynamic gait features such as step length, walk speed, joint angles and inter-joint distances, along with various statistics (mean, standard deviation or maximum) of their signals. These, in particular, are the horizontal and vertical distances of selected joint pairs by Ahmed~\etal~\cite{AAS14}, lower limb triangles by Ali~\etal~\cite{AWLSWZ16}, lower body angles, step length, cycle time and velocity by Andersson and Araujo~\cite{AA15}, lower limb (hips, knees and ankles) angles by Ball~\etal~\cite{BRRV12}, axis rotations of the major bones by Kwolek~\etal~\cite{KKMJ14}, eleven static body parameters, and step length and walk speed by Preis~\etal~\cite{PKWL12}. Dikovski~\etal~\cite{DMG14} construct seven different feature sets from a broad spectrum of geometric features, such as static body parameters, joint angles and inter-joint distances aggregated within a gait cycle, along with various statistics. Sinha~\etal~\cite{SCB13} combine areas of upper and lower body, inter-joint distances as well as all of the features introduced by Ball~\etal~\cite{BRRV12} and Preis~\etal~\cite{PKWL12}. These features are convenient for visualizations and for intuitive understanding, but their schematism and human-interpretability are unnecessary for automatic identification.

Instead, we~\cite{BS16a,BS16b} prefer to statistically learn the features on an auxiliary labeled database with the goal of maximally separating the identity classes in the feature space and use these features to identify all potential walkers. Their linear model is learned in a supervised manner through (1)~a~modification of Fisher's Linear Discriminant Analysis~\cite{F36} with Maximum Margin Criterion (MMC) and (2)~a~combination of Principal Component Analysis and Linear Discriminant Analysis (PCA+LDA) to project the high-dimensional input data onto low-dimensional sub-spaces. The similarity of templates is expressed in the Mahalanobis distance function. Both these and the non-learning approaches create an unsupervised environment suitable for searching for similar templates and for clustering templates into potential identities, which is the main focus of the application outlined in the following phase.

\subsection*{Phase IV -- Identifying Walkers}

Identification is most commonly formulated as a classification problem: A walker's identity is established (classified) by picking one from the pool of registered identities. This model is suitable for applications where participants reveal their identities at registration (closed-set identification). During video surveillance, on the other hand, new identities can appear on the fly (open-set identification), and labeled data for all the people encountered may not always be available. Nobody is claiming their identities since people are recorded without their consent. Needless to say, this task requires a different understanding of what a person's identity~is.

\begin{figure}[tb]
\centering
\includegraphics[width=0.28\textwidth]{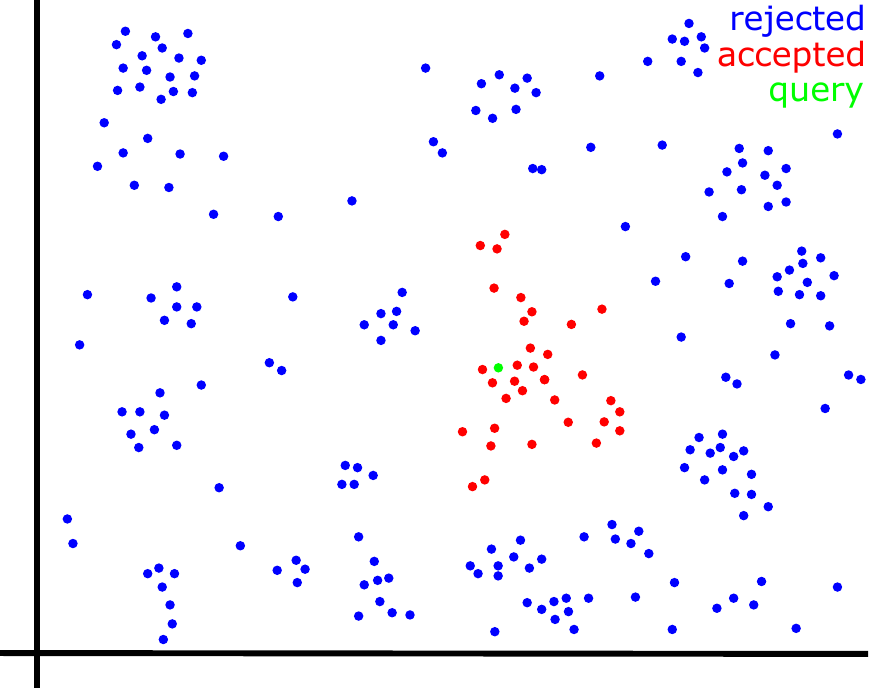}
\includegraphics[width=0.47\textwidth]{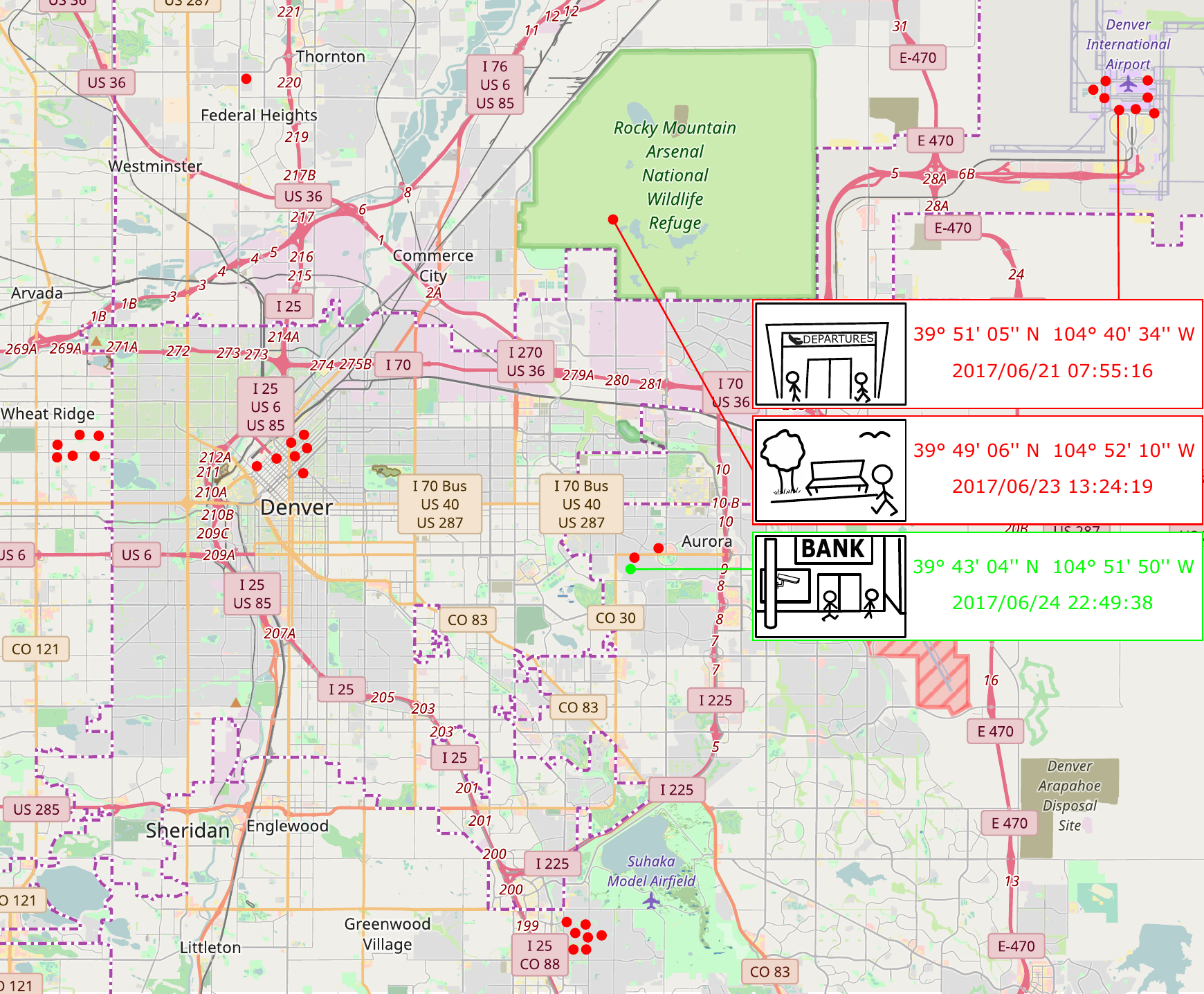}
\caption{Illustration of a person's location trace. Each point represents a gait template and contains additional information about the place and time of the corresponding incident within the surveillance system. Given a query template (green dot), the retrieved cluster of templates (red dots) is expected to contain other templates of the same person. Map obtained from \protect\url{https://www.openstreetmap.org}.}
\label{f2}
\end{figure}

You are how you walk. Your identity is your gait pattern itself. Instead of classifying walker identities as names or numbers that are not available in any case, a forensic investigator rather asks for information about their appearances captured by surveillance system -- their location trace (see Figure~\ref{f2}) that includes timestamp and geolocation of each appearance. In the suggested application, walkers are clustered rather than classified. Identification is carried out as a query-by-example: A similarity search query ranks recorded gait templates on the basis of their similarity to the query template, which represents their likelihood of belonging to the same person. A decision mechanism operates on the basis of clustering or thresholding to determine which of the templates are finally accepted to establish the location trace. Although one can allow more templates to be accepted, doing so will enlarge the location trace with a chance of falsely accepting some templates of another person.

\section{Evaluation}
\label{eval}

We have based our evaluation on the gait recognition framework~\cite{BS16c}. The evaluation focuses on the Phase III in Section~\ref{flow}. We investigate the impact of externalizing the learning identities as a resolution for walker uncooperativeness by evaluating \emph{discriminativeness} of the feature space learned by a certain amount of separate learning data. One also desires a system with a model that does not collapse even if data are mapped wrongly due to inaccuracies or failures of the data acquisition technology, for which we evaluate \emph{robustness} incomplete and noisy data. The final stage is the evaluation \emph{clusterability} of individual feature spaces for potential data pre-processing.

We implemented and evaluated all competitive MoCap gait identification methods~\cite{AAS14,AWLSWZ16,AA15,BS16a,BRRV12,DMG14,KKMJ14,PKWL12,SCB13}. The state-of-the-art records additional methods~\cite{APG15,JWZS15,KSKJW14,NV12,SVBZ12} which we have implemented but not evaluated due to the high demands they place on computational resources.

For the purpose of evaluation, we selected the MoCap database of the CMU Graphics Lab~\cite{CMU03}, which is available under the Creative Commons license. Normalization and extraction of gait cycles from this database is described in~\cite{BS16c} and is available for download at~\cite{WWW}. The gait data hold $\gC=64$~walking subjects that performed $\gN=5\,923$~samples in total, which resulted in an average of about $83$~samples per subject. Data in the measurement (sample) space have the form $\gG=\left\{\left(\gGn{n},\gLAMBDAn{n}\right)\right\}_{n=1}^\gN$ where $\gGn{n}$ is a tensorial representation of a gait sample (a single gait cycle) that contains 3D spatial coordinates (positions) of joints in all video frames, normalized with respect to the person's position and direction of walking. Each of the $\gN$ learning samples falls strictly into one of the $\gC$ identity classes representing a single walker labeled $\gLAMBDAn{n}$. A~class $\gCLASc{c}\subseteq\gG$ has $\gNc{c}$ samples. Classes $\gCLAS=\left\{\gCLASc{c}\right\}_{c=1}^\gC$ are complete and mutually exclusive. We say that samples $\left(\gGn{n},\gLAMBDAn{n}\right)$ and $\left(\gGn{n'},\gLAMBDAn{n'}\right)$ share a common walker if and only if they belong to the same class: $\left(\gGn{n},\gLAMBDAn{n}\right),\left(\gGn{n'},\gLAMBDAn{n'}\right)\in\gCLASc{c}\Leftrightarrow\gLAMBDAn{n}=\gLAMBDAn{n'}$.

We evaluate the implemented methods with the cross-identity evaluation setup, in which the collections of learning data $\gL{\gG}=\left\{\left(\gGn{n},\gLAMBDAn{n}\right)\right\}_{n=1}^{\gL{\gN}}$ of $\gL{\gC}$ identities and evaluation data $\gE{\gG}=\left\{\left(\gGn{n},\gLAMBDAn{n}\right)\right\}_{n=1}^{\gE{\gN}}$ of $\gE{\gC}$ identities are disjunct. An evaluation configuration is parametrized by $\left(\gL{\gC},\gE{\gC}\right)$ specifying how many learning and how many evaluation identity classes are selected from the database.

Models are eventually learned on the learning part and evaluated on the evaluation part transformed into individual feature spaces $\gE{\gG}=\left\{\left(\gGn{n},\gLAMBDAn{n}\right)\right\}_{n=1}^{\gE{\gN}}\rightarrow\gE{\gH{\gG}}=\left\{\gGnH{n},\gLAMBDAn{n}\right\}_{n=1}^{\gE{\gN}}$ of $\gE{\gC}$ identities, as determined by corresponding methods. Evaluation results are presented in terms of the following metrics:
\def\ditem#1{\item[\textbullet\:\normalfont\textit{#1}\!]}
\begin{description}[style=unboxed,leftmargin=0cm]
\setlength{\parskip}{0pt}
\ditem{Davies-Bouldin Index:} $\DBI=\frac{1}{\gE{\gC}}\sum_{c=1}^{\gE{\gC}}\max\limits_{1 \leq c' \leq \gC,\,c' \neq c}\frac{\sigma_c+\sigma_{c'}}{\gDELTAccH{\gMcH{c}}{\gMcH{c'}}}$\\
where $\sigma_c=\frac{1}{\gNc{c}}\sum_{n=1}^{\gNc{c}}\gDELTAccH{\gGnH{n}}{\gMcH{c}}$ is the average distance of all elements in identity class $\gCLASc{c}$ to its centroid, and analogically for $\sigma_{c'}$. Templates of low intra-class distances and of high inter-class distances have a low \DBI.
\ditem{Silhouette Coefficient:} $\SC=\frac{1}{\gE{\gN}}\sum_{n=1}^{\gE{\gN}}\frac{b(\gGnH{n})-a(\gGnH{n})}{\max\left\{a\left(\gGnH{n}),b(\gGnH{n}\right)\right\}}$\\
where $a(\gGnH{n})=\frac{1}{\gNc{c}}\sum_{n'=1}^{\gNc{c}}\gDELTAccH{\gGnH{n}}{\gGnH{n'}}$ is the average distance from $\gGnH{n}$ to other samples within the same identity class and $b(\gGnH{n})=\min\limits_{1 \leq c' \leq \gC,\,c' \neq c}\frac{1}{\gNc{c'}}\sum_{n'=1}^{\gNc{c'}}\gDELTAccH{\gGnH{n}}{\gGnH{n'}}$ is the average distance of $\gGnH{n}$ to the samples in the closest class. It is clear that $-1\leq\SC\leq1$ and a \SC close to one means that classes are appropriately separated.
\ditem{area under Receiver Operating Characteristic curve (\ROC)}
\ditem{area under Precision-Recall curve (\PR)}
\end{description}

The system can potentially employ various clustering mechanisms that need a high degree of accuracy. Consider a clustering algorithm that returns the clusters $\gCLUS=\left\{\gCLUSc{c'}\right\}_{c'=1}^\gC$ approximating the real identity classes $\gCLAS=\left\{\gCLASc{c}\right\}_{c=1}^\gC$. The following metrics evaluate a clustering algorithm together with a particular feature extraction method, which gives an insight into clusterability of the corresponding feature space:
\begin{description}[style=unboxed,leftmargin=0cm]
\setlength{\parskip}{0pt}
\ditem{Purity:} $\PURITY=\frac{1}{\gE{\gN}}\sum_{\gCLUSc{c'}\in\gCLUS}\max_{\gCLASc{c}\in\gCLAS}\left|\gCLUSc{c'}\cap\gCLASc{c}\right|$
\ditem{Rand Index:} $\RI=\frac{\TP+\TN}{\TP+\FP+\FN+\TN}$
\ditem{F-measure:} $\F=\frac{2pr}{p+r}$ where $p=\frac{\TP}{\TP+\FP}$ and $r=\frac{\TP}{\TP+\FN}$
\ditem{Jaccard Index:} $\JI=\frac{\TP}{\TP+\FP+\FN}$
\ditem{Fowlkes-Mallows Index:} $\FMI=\sqrt{\frac{\TP}{\TP+\FP}\cdot\frac{\TP}{\TP+\FN}}$\\
where \TP is the number of true positives, \TN of true negatives, \FP of false positives, and \FN of false negatives. A~pair of templates of the same identity falling in the same cluster is a true positive, of different identities in different clusters is a true negative, of different identities in the same cluster is a false positive, and of the same identity in different clusters is a false negative.
\end{description}

\begin{figure*}[tb]
\centering
\begin{subfigure}{0.7\textwidth}
  \includegraphics[width=\textwidth]{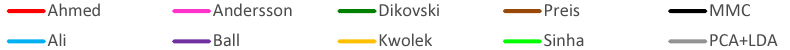}
\end{subfigure}
\begin{subfigure}{0.49\textwidth}
  \medskip
  \includegraphics[width=\textwidth]{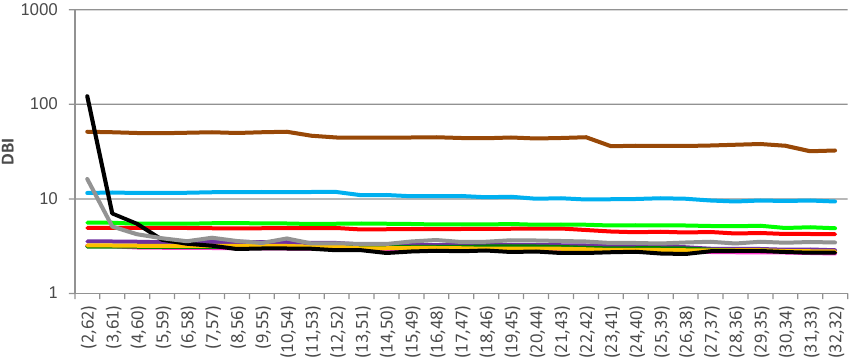}
  \caption{\DBI}
\end{subfigure}
\begin{subfigure}{0.49\textwidth}
  \medskip
  \includegraphics[width=\textwidth]{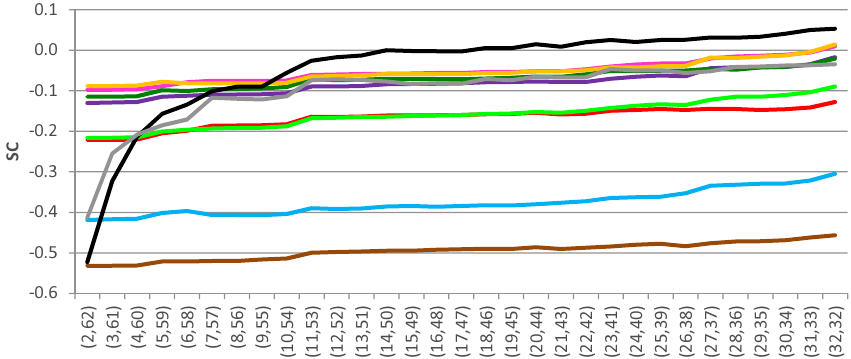}
  \caption{\SC}
\end{subfigure}
\begin{subfigure}{0.49\textwidth}
  \medskip
  \includegraphics[width=\textwidth]{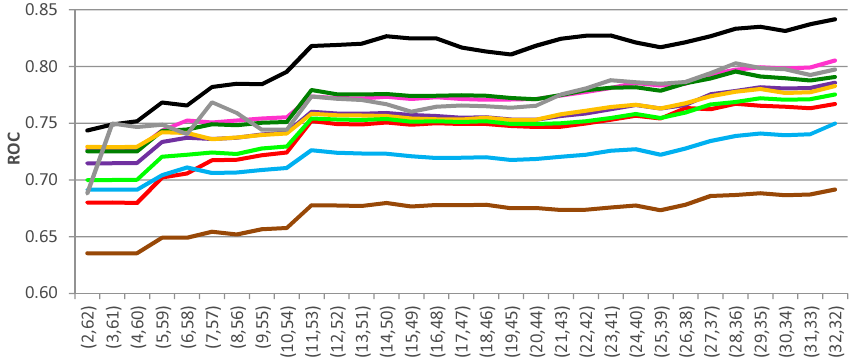}
  \caption{\ROC}
\end{subfigure}
\begin{subfigure}{0.49\textwidth}
  \medskip
  \includegraphics[width=\textwidth]{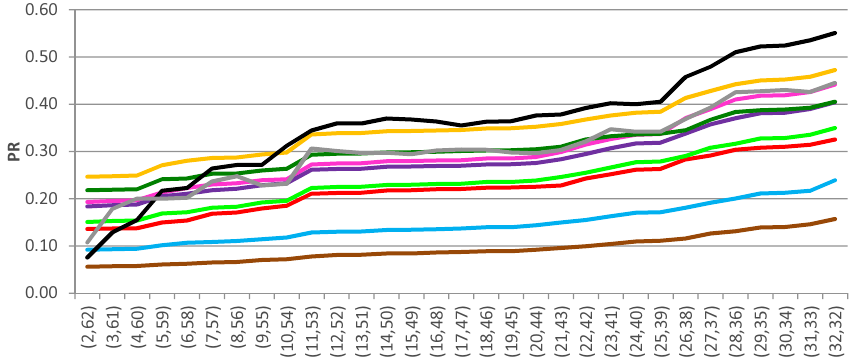}
  \caption{\PR}
\end{subfigure}
\vspace{-3pt}%
\caption{Simulations with 31 different $\left(\gL{\gC},\gE{\gC}\right)$ configurations (horizontal axes) on four evaluation metrics (vertical axes).}
\label{f3}
\medskip
\includegraphics[width=\textwidth]{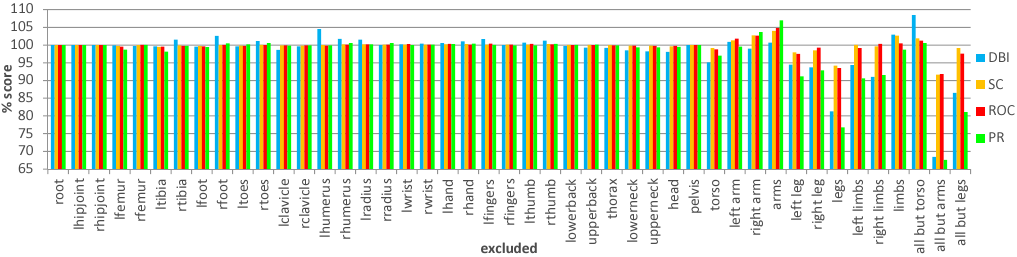}
\vspace{-16pt}%
\caption{Four evaluated metrics of the MMC method with incomplete data. In each column a subset of joints is systematically excluded from the input: first 31~columns (\joint{root} -- \joint{head}) exclude a single joint and last 14~columns (\joint{pelvis} -- \joint{all but legs}) exclude multiple joints. Structure of the human body is the following:
\joint{head},
$\joint{pelvis}=\left\{\joint{root},\joint{lhipjoint},\joint{rhipjoint}\right\}$, $\joint{left leg}=\left\{\joint{lfemur},\joint{ltibia},\joint{lfoot},\joint{ltoes}\right\}$, $\joint{left arm}=\left\{\joint{lhumerus},\joint{lradius},\joint{lwrist},\joint{lhand},\joint{lfingers},\joint{lthumb}\right\}$, $\joint{right leg}=\left\{\joint{rfemur},\joint{rtibia},\joint{rfoot},\joint{rtoes}\right\}$, $\joint{right arm}=\left\{\joint{rhumerus},\joint{rradius},\joint{rwrist},\joint{rhand},\joint{rfingers},\joint{rthumb}\right\}$, $\joint{torso}=\left\{\joint{lowerback},\joint{upperback},\joint{thorax},\joint{lowerneck},\joint{upperneck},\joint{lclavicle},\joint{rclavicle}\right\}$. Configuration $\left(9,55\right)$.}
\label{f4}
\medskip
\begin{subfigure}{0.49\textwidth}
  \captionsetup{width=0.97\textwidth}
  \includegraphics[width=0.49\textwidth]{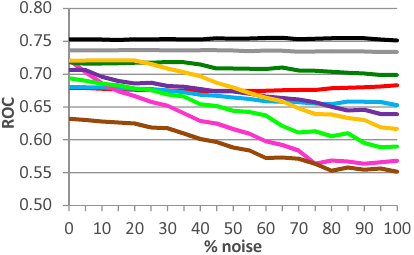}
  \includegraphics[width=0.49\textwidth]{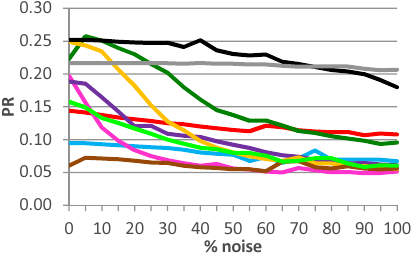}
  \caption{\ROC and \PR for an $x$\% noise simulated by multiplying each measured value by a random number in interval $\left(1-\nicefrac{x}{100},1+\nicefrac{x}{100}\right)$.}
\end{subfigure}
\begin{subfigure}{0.49\textwidth}
  \captionsetup{width=0.97\textwidth}
  \includegraphics[width=0.49\textwidth]{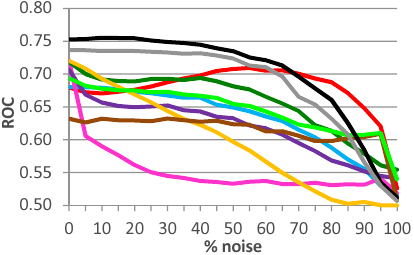}
  \includegraphics[width=0.49\textwidth]{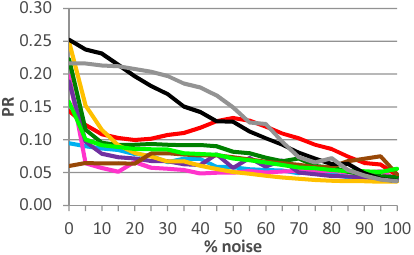}
  \caption{\ROC and \PR for an $x$\% noise simulated by substituting each measured value with a random value with probability $x$\%.}
\end{subfigure}
\vspace{-3pt}%
\caption{\ROC and \PR of all implemented methods with noisy data. Configuration $\left(9,55\right)$.}
\label{f5}
\end{figure*}

\paragraph{Discriminativeness}
Designing a surveillance system with uncooperative subjects, one should also consider the availability of an auxiliary learning database that is large enough to be used for learning the model. In the following series of experiments we measure the four evaluation metrics on a sequence of configurations with an increasing number of learning identities and the rest of the database as an
evaluation part. Given that the benchmark dataset contains 64~identities in total, these configurations range from $\left(2,62\right)$ to $\left(32,32\right)$. Technically, it is possible to continue up until $\left(62,2\right)$, but configurations with more learning than evaluation identities are insignificant. Each configuration $\left(\gL{\gC},64-\gL{\gC}\right)$ is constructed from the previous configuration $\left(\gL{\gC}-1,64-\gL{\gC}\right)+1$ by picking one identity in the evaluation part at random and moving it into the learning part. Observing from Figure~\ref{f4}, the discriminativeness of the approaches based on statistical learning (MMC and PCA+LDA) grows quickly on the first configurations with very few learning identities, which one can interpret as an analogy to the Pareto (80--20) principle. The results indicate that even 10~identities can be enough for learning the MMC transform to identify 54~other people more accurately than the other methods tested. Roughly speaking, the MMC method achieves the top results in all of the metrics at a configuration of about $\left(10,54\right)$ and keeps or increases its discriminativeness further on. This experiment provides a lower bound estimate for the volume of learning data given the volume of the population for surveillance: With a learning database smaller than this lower bound, taking the non-learned geometric features of Dikovski~\etal~\cite{DMG14} or Kwolek~\etal~\cite{KKMJ14} is recommended, otherwise the MMC method best identifies walkers within a population of more than triple the learning identities.

\paragraph{Robustness}
Video surveillance environments are vulnerable to the accuracy of data acquisition. We conducted two series of experiments where we assumed a $\left(9,55\right)$ configuration with fixed learning and evaluation identities on which we simulated measurement errors by (1)~excluding some data and by (2)~adding random noise. Incomplete data are simulated by excluding various subsets of joints with all their tracked positions and score percentages of incomplete data (with subscript \new) to complete data (with subscript \old) were calculated for \DBI as $\nicefrac{100\cdot\DBI_\old}{\DBI_\new}$, for \SC~as $\nicefrac{100\cdot(\SC_\new+1)}{\SC_\old+1}$, for \ROC as $\nicefrac{100\cdot\ROC_\new}{\ROC_\old}$, and for \PR as $\nicefrac{100\cdot\PR_\new}{\PR_\old}$. A~noise of $x$\% was simulated in two ways: by (2a)~multiplying each measured value by a random number from the interval $\left(1-\nicefrac{x}{100},1+\nicefrac{x}{100}\right)$ and by (2b)~substituting each measured value with a random value between 0 and 1 with $x$\% probability in which the 100\% noise represents completely random data. All methods were evaluated on noisy data but only the MMC method was evaluated on incomplete data as other methods do not give instructions for dealing with incomplete data. Figures~\ref{f4} and~\ref{f5} illustrate how the methods cope with incomplete and noisy data, respectively. As for the data incompleteness, one can observe that particular joints are more discriminatory than others as their exclusion causes a more significant drop in scores. This information can be potentially used for fine-tuning the hand-picked geometric features. Furthermore, input data pruning has a positive impact on the duration of model learning. Regarding the noisy data, note that the first type of noise is shrouding less than the second type as the measured values change only up to double and they preserve their means and covariances. The results make it clear that the methods of geometric features drop in score more quickly than the MMC and PCA+LDA methods.

\paragraph{Clusterability}
Various pre-clustering techniques can be applied to the feature space in order to improve the accuracy of the location trace retrieval. One perfectly accurate location trace would be the cluster of gait templates of and only of the query walker, although this request is near impossible on large datasets. The last experiment takes again the $\left(9,55\right)$ configuration and measures the purity and other four indexes of the clustering obtained by K-Means into $\mathrm{K}=\gE{\gC}=55$ clusters. Results in Table~\ref{t1} show that the geometric features of Andersson~\etal~\cite{AA15}, Dikovski~\etal~\cite{DMG14}, Kwolek~\etal~\cite{KKMJ14} and the latent features learned by MMC and PCA+LDA define the most suitable sub-spaces for K-Means clustering.

\begin{table}[tb]
\caption{Methods evaluated on the five clusterability metrics. Configuration $\left(9,55\right)$.}
\vspace{-10pt}%
\centering\tabcolsep4.5pt
\begin{tabular}{r|ccccc}
\toprule[1pt]
method	& \PURITY	& \RI	& F	& JI	& FMI \\
\midrule[0.4pt]
Ahmed	& 0.3402	& 0.9486	& 0.1388	& 0.0746	& 0.1418 \\
Ali	& 0.2677	& 0.9472	& 0.0986	& 0.0519	& 0.1013 \\
Andersson	& 0.3574	& 0.9526	& 0.26\hphantom{00}	& 0.1494	& 0.262\hphantom{0} \\
Ball	& 0.3409	& 0.9491	& 0.1581	& 0.0859	& 0.1611 \\
Dikovski	& 0.4446	& 0.9542	& 0.2583	& 0.1483	& 0.2619 \\
Kwolek	& 0.4571	& 0.9496	& 0.2319	& 0.1311	& 0.2329 \\
Preis	& 0.1778	& 0.9464	& 0.0462	& 0.0237	& 0.0482 \\
Sinha	& 0.3069	& 0.9473	& 0.1143	& 0.0606	& 0.1169 \\
MMC	& 0.4491	& 0.9538	& 0.2147	& 0.1203	& 0.2202 \\
PCA+LDA	& 0.437\hphantom{0}	& 0.9538	& 0.224\hphantom{0}	& 0.1261	& 0.2291 \\
\bottomrule[1pt]
\end{tabular}
\label{t1}
\vspace{-3pt}%
\end{table}

\section{Conclusion}
\label{concl}

We present a plausible high-level workflow of a MoCap gait identification system in a video surveillance environment with uncooperative walkers. The main focus is on the evaluation of whether the Phase~III of the workflow (extracting gait features) can meet the requirements of the Phase~IV of the workflow (a target application for identifying walkers). Eight methods for extracting geometric gait features and two methods for statistical learning the features have been implemented and evaluated on the CMU MoCap database of 64~subjects and 5,923~gait samples. Feature space of each method is evaluated for (1)~discriminativeness, (2)~robustness to incomplete and noisy data and (3)~clusterability. With the evaluation set of 55 identities, the MMC method learned on 9 identities achieves the top results in discriminativeness and improves with increasing learning identities; together with the PCA+LDA method they are highly robust to noisy and incomplete data; and result in a rather pure clustering. The MMC method appears to be learning the top quality features, reaching the state-of-the-art in MoCap gait identification.

Our suggested concept of person identification is based on a completely different perspective from previous approaches: According to the phrase \emph{You are how you walk}, a walker identity is represented as their location trace, that is, a cluster of gait templates from their incidents within the surveillance system. This concept allows for data-driven models that (1)~can be learned from different people as well as from different scenes and datasets, making it more generally applicable with limited data per person in a learning set, even if there is only a single gait sample available for each person, and (2)~do not make assumptions about the learning and evaluation sets having the same covariate conditions as long as the auxiliary learning set is rich in them to identify as many people as possible. This makes it particularly suitable for applications of uncooperative recognition, such as walker re-identification~\cite{MZGXHLZ16,ZYH16} or next location prediction~\cite{MC16,MSPC16}.

\paragraph{Acknowledgments}
Data used in this project was created with funding from NSF EIA-0196217 and was obtained from \url{http://mocap.cs.cmu.edu}~\cite{CMU03}. Our extracted database and evaluation framework are available online at \url{https://gait.fi.muni.cz} to support reproducibility of results.

{\small
\bibliographystyle{ieee}
\bibliography{ref}

\begin{thebibliography}{10}\itemsep=-1pt

\bibitem{APG15}
F.~Ahmed, P.~P. Paul, and M.~L. Gavrilova.
\newblock {DTW-Based Kernel and Rank-Level Fusion for 3D Gait Recognition Using
  Kinect}.
\newblock {\em The Visual Computer}, 31(6):915--924, 2015.

\bibitem{AAS14}
M.~Ahmed, N.~Al-Jawad, and A.~Sabir.
\newblock {Gait Recognition Based on Kinect Sensor}.
\newblock {\em Proc. SPIE}, 9139:91390B--91390B--10, 2014.

\bibitem{AWLSWZ16}
S.~Ali, Z.~Wu, X.~Li, N.~Saeed, D.~Wang, and M.~Zhou.
\newblock {\em {Transactions on Computational Science XXVI: Special Issue on
  Cyberworlds and Cybersecurity}}, chapter {Applying Geometric Function on
  Sensors 3D Gait Data for Human Identification}, pages 125--141.
\newblock Springer, Berlin, Heidelberg, 2016.

\bibitem{AA15}
V.~O. Andersson and R.~M. Araujo.
\newblock {Person Identification Using Anthropometric and Gait Data from Kinect
  Sensor}.
\newblock In {\em {Proc. of the Twenty-Ninth AAAI Conference on Artificial
  Intelligence (AAAI-15)}}, pages 425--431. AAAI Press, 2015.

\bibitem{AMAMR15}
E.~Auvinet, F.~Multon, C.-E. Aubin, J.~Meunier, and M.~Raison.
\newblock Detection of gait cycles in treadmill walking using a {Kinect}.
\newblock {\em Gait \& Posture}, 41(2):722--725, 2015.

\bibitem{BS16a}
M.~Balazia and P.~Sojka.
\newblock {Learning Robust Features for Gait Recognition by Maximum Margin
  Criterion}.
\newblock In {\em {Proc. of 23rd International Conference on Pattern
  Recognition, ICPR 2016}}, pages 901--906. IEEE, 2016.

\bibitem{BS16b}
M.~Balazia and P.~Sojka.
\newblock Walker-independent features for gait recognition from motion capture
  data.
\newblock In A.~Robles-Kelly, M.~Loog, B.~Biggio, F.~Escolano, and R.~Wilson,
  editors, {\em Structural, Syntactic, and Statistical Pattern Recognition:
  Joint IAPR International Workshop, S+SSPR 2016, M{\'e}rida, Mexico, November
  29--December 2, 2016, Proceedings}, pages 310--321, Cham, 2016. Springer
  International Publishing.

\bibitem{BS16c}
M.~Balazia and P.~Sojka.
\newblock An evaluation framework and database for mocap-based gait recognition
  methods.
\newblock In B.~Kerautret, M.~Colom, and P.~Monasse, editors, {\em Reproducible
  Research in Pattern Recognition: First International Workshop, RRPR 2016,
  Canc{\'u}n, Mexico, December 4, 2016, Revised Selected Papers}, pages 33--47,
  Cham, 2017. Springer International Publishing.

\bibitem{WWW}
M.~Balazia and P.~Sojka.
\newblock Gait recognition from motion capture data, 2017.
\newblock \url{https://gait.fi.muni.cz}.

\bibitem{BRRV12}
A.~Ball, D.~Rye, F.~Ramos, and M.~Velonaki.
\newblock Unsupervised clustering of people from 'skeleton' data.
\newblock In {\em Proceedings of the Seventh Annual ACM/IEEE International
  Conference on Human-Robot Interaction}, HRI '12, pages 225--226, New York,
  NY, USA, 2012. ACM.

\bibitem{CX16}
X.~Chen and J.~Xu.
\newblock Uncooperative gait recognition: Re-ranking based on sparse coding and
  multi-view hypergraph learning.
\newblock {\em Pattern Recognition}, 53:116 -- 129, 2016.

\bibitem{CCH12}
W.~Choensawat, W.~Choi, and K.~Hachimura.
\newblock Similarity retrieval of motion capture data based on derivative
  features.
\newblock {\em Advanced Computational Intelligence and Intelligent
  Informatics}, 16(1):13--23, 2012.

\bibitem{CMU03}
{CMU Graphics Lab}.
\newblock {Carnegie-Mellon Motion Capture (MoCap) Database}, 2003.
\newblock \url{http://mocap.cs.cmu.edu}.

\bibitem{DMG14}
B.~Dikovski, G.~Madjarov, and D.~Gjorgjevikj.
\newblock {Evaluation of Different Feature Sets for Gait Recognition Using
  Skeletal Data from Kinect}.
\newblock In {\em 37th Intl.\ Convention on Information and Communication
  Technology, Electronics and Microelectronics}, pages 1304--1308, May 2014.

\bibitem{DF16}
M.~Ding and G.~Fan.
\newblock Articulated and generalized gaussian kernel correlation for human
  pose estimation.
\newblock {\em IEEE Transactions on Image Processing}, 25(2):776--789, Feb
  2016.

\bibitem{F36}
R.~A. Fisher.
\newblock {The Use of Multiple Measurements in Taxonomic Problems}.
\newblock {\em Annals of Eugenics}, 7(2):179--188, 1936.

\bibitem{HALP13}
S.~Han, M.~Achar, S.~Lee, and F.~Pe{\~{n}}a-Mora.
\newblock Empirical assessment of a rgb-d sensor on motion capture and action
  recognition for construction worker monitoring.
\newblock {\em Visualization in Engineering}, 1(1):6, 2013.

\bibitem{HPLAYL16}
A.~Haque, B.~Peng, Z.~Luo, A.~Alahi, S.~Yeung, and F.~Li.
\newblock Viewpoint invariant 3d human pose estimation with recurrent error
  feedback.
\newblock {\em CoRR}, abs/1603.07076, 2016.

\bibitem{HCCCLW15}
M.~C. Hu, C.~W. Chen, W.~H. Cheng, C.~H. Chang, J.~H. Lai, and J.~L. Wu.
\newblock {Real-Time Human Movement Retrieval and Assessment With Kinect
  Sensor}.
\newblock {\em IEEE Transactions on Cybernetics}, 45(4):742--753, Apr. 2015.

\bibitem{JWZS15}
S.~Jiang, Y.~Wang, Y.~Zhang, and J.~Sun.
\newblock {Real Time Gait Recognition System Based on Kinect Skeleton Feature}.
\newblock In C.~Jawahar and S.~Shan, editors, {\em Computer Vision -- ACCV 2014
  Workshops}, volume 9008 of {\em LNCS}, pages 46--57. Springer, 2015.

\bibitem{KN14}
I.~Kapsouras and N.~Nikolaidis.
\newblock Action recognition in motion capture data using a bag of postures
  approach.
\newblock In {\em Pattern Recognition (ICPR), 2014 22nd International
  Conference on}, pages 2649--2654, Aug. 2014.

\bibitem{KSKJW14}
T.~Krzeszowski, A.~Switonski, B.~Kwolek, H.~Josinski, and K.~Wojciechowski.
\newblock {DTW-Based Gait Recognition from Recovered 3-D Joint Angles and
  Inter-ankle Distance}.
\newblock In L.~J. Chmielewski, R.~Kozera, B.-S. Shin, and K.~Wojciechowski,
  editors, {\em Proc. of Computer Vision and Graphics: International
  Conference, ICCVG 2014, Warsaw, Poland}, volume 8671 of {\em LNCS}, pages
  356--363. Springer, Sept. 2014.

\bibitem{NV12}
M.~S.~N. Kumar and R.~V. Babu.
\newblock Human gait recognition using depth camera: A covariance based
  approach.
\newblock In {\em Proc. of the Eighth Indian Conference on Computer Vision,
  Graphics and Image Processing}, ICVGIP '12, pages 20:1--20:6, New York, NY,
  USA, 2012. ACM.

\bibitem{KKMJ14}
B.~Kwolek, T.~Krzeszowski, A.~Michalczuk, and H.~Josinski.
\newblock {3D Gait Recognition Using Spatio-Temporal Motion Descriptors}.
\newblock In {\em Proc. of Intelligent Information and Database Systems: 6th
  Asian Conference, ACIIDS 2014, Bangkok, Thailand, Part II}, volume 8398 of
  {\em LNCS}, pages 595--604. Springer, Apr. 2014.

\bibitem{LLMYD14}
D.~Leightley, B.~Li, J.~S. McPhee, M.~H. Yap, and J.~Darby.
\newblock {\em Exemplar-Based Human Action Recognition with Template Matching
  from a Stream of Motion Capture}, pages 12--20.
\newblock Springer International Publishing, Cham, 2014.

\bibitem{MZGXHLZ16}
X.~Ma, X.~Zhu, S.~Gong, X.~Xie, J.~Hu, K.-M. Lam, and Y.~Zhong.
\newblock Person re-identification by unsupervised video matching.
\newblock {\em Pattern Recogn.}, 65(C):197--210, May 2017.

\bibitem{MC16}
U.~Mahbub and R.~Chellappa.
\newblock {PATH:} person authentication using trace histories.
\newblock {\em CoRR}, abs/1610.07935, 2016.

\bibitem{MSPC16}
U.~Mahbub, S.~Sarkar, V.~M. Patel, and R.~Chellappa.
\newblock Active user authentication for smartphones: A challenge data set and
  benchmark results.
\newblock In {\em 2016 IEEE 8th International Conference on Biometrics Theory,
  Applications and Systems (BTAS)}, pages 1--8, Sept 2016.

\bibitem{MX14}
R.~Martín-Félez and T.~Xiang.
\newblock Uncooperative gait recognition by learning to rank.
\newblock {\em Pattern Recognition}, 47(12):3793--3806, 2014.

\bibitem{PKWL12}
J.~Preis, M.~Kessel, M.~Werner, and C.~Linnhoff-Popien.
\newblock {Gait Recognition with Kinect}.
\newblock In {\em 1st International Workshop on Kinect in Pervasive Computing,
  New Castle, UK, June 18--22}, pages 1--4, 2012.

\bibitem{SVBZ12}
J.~Sedmidubsky, J.~Valcik, M.~Balazia, and P.~Zezula.
\newblock {Gait Recognition Based on Normalized Walk Cycles}.
\newblock In {\em Advances in Visual Computing}, volume 7432 of {\em LNCS},
  pages 11--20. Springer, 2012.

\bibitem{SCB13}
A.~Sinha, K.~Chakravarty, and B.~Bhowmick.
\newblock {Person Identification Using Skeleton Information from Kinect}.
\newblock In {\em ACHI 2013: Proc. of the Sixth Intl. Conf. on Advances in
  CHI}, pages 101--108, 2013.

\bibitem{VSBZ12}
J.~Valcik, J.~Sedmidubsky, M.~Balazia, and P.~Zezula.
\newblock Identifying walk cycles for human recognition.
\newblock In M.~Chau, A.~G. Wang, W.~T. Yue, and H.~Chen, editors, {\em
  Intelligence and Security Informatics: Pacific Asia Workshop, PAISI 2012,
  Kuala Lumpur, Malaysia, May 29, 2012. Proceedings}, pages 127--135, Berlin,
  Heidelberg, 2012. Springer Berlin Heidelberg.

\bibitem{VR14}
S.~Vantigodi and V.~B. Radhakrishnan.
\newblock Action recognition from motion capture data using meta-cognitive rbf
  network classifier.
\newblock In {\em Intelligent Sensors, Sensor Networks and Information
  Processing (ISSNIP), 2014 IEEE Ninth International Conference on}, pages
  1--6, Apr. 2014.

\bibitem{ZYH16}
L.~Zheng, Y.~Yang, and A.~G. Hauptmann.
\newblock Person re-identification: Past, present and future.
\newblock {\em arXiv preprint arXiv:1610.02984}, 2016.

\end{thebibliography}
}

\end{document}